\documentclass{article}

\PassOptionsToPackage{numbers, compress}{natbib}
\usepackage[preprint]{neurips_2020}

\usepackage[utf8]{inputenc} % allow utf-8 input
\usepackage[T1]{fontenc}    % use 8-bit T1 fonts
\usepackage[colorlinks]{hyperref}       % hyperlinks (color)
\usepackage{url}            % simple URL typesetting
\usepackage{booktabs}       % professional-quality tables
\usepackage{amsfonts}       % blackboard math symbols
\usepackage{nicefrac}       % compact symbols for 1/2, etc.
\usepackage{microtype}      % microtypography
\usepackage{textcomp}
\usepackage{graphicx}
\usepackage{multirow}
\usepackage{xcolor}
\usepackage{wrapfig}
\usepackage{amsmath}
\usepackage{booktabs}
\usepackage{caption}
\usepackage{subcaption}

\title{Few-shot Visual Reasoning with\\ Meta-analogical Contrastive Learning}

\author{%
 Youngsung Kim \\
 Samsung Advanced Institute of Technology \\
 \texttt{yskim.ee@gmail.com} \\
 \AND
 Jinwoo Shin \hspace{0.6in}
 Eunho Yang  \hspace{0.6in}
 Sung Ju Hwang  \\
Korea Advanced Institute of Science and Technology (KAIST)\\
\texttt{\{jinwoos,eunhoy,sjhwang82\}@kaist.ac.kr}
}
\begin{document}

\maketitle

\begin{abstract}
While humans can solve a visual puzzle that requires logical reasoning by observing only few samples, it would require training over large amount of data for state-of-the-art deep reasoning models to obtain similar performance on the same task. In this work, we propose to solve such a few-shot (or low-shot) visual reasoning problem, by resorting to \emph{analogical reasoning}, which is a unique human ability to identify structural or relational similarity between two sets. Specifically, given training and test sets that contain the same type of visual reasoning problems, we extract the structural relationships between elements in both domains, and enforce them to be as similar as possible with analogical learning. We repeatedly apply this process with slightly modified queries of the same problem under the assumption that it does not affect the relationship between a training and a test sample. This allows to learn the relational similarity between the two samples in an effective manner even with a single pair of samples. We validate our method on RAVEN dataset, on which it outperforms state-of-the-art method, with larger gains when the training data is scarce. We further meta-learn our analogical contrastive learning model over the same tasks with diverse attributes, and show that it generalizes to the same visual reasoning problem with unseen attributes. 
\end{abstract}

\section{Introduction}
\begin{figure}
%\vspace{-0.7cm}
     \centering
     \begin{subfigure}[b]{0.37\textwidth}
         \centering
        \includegraphics[width=0.9\linewidth]{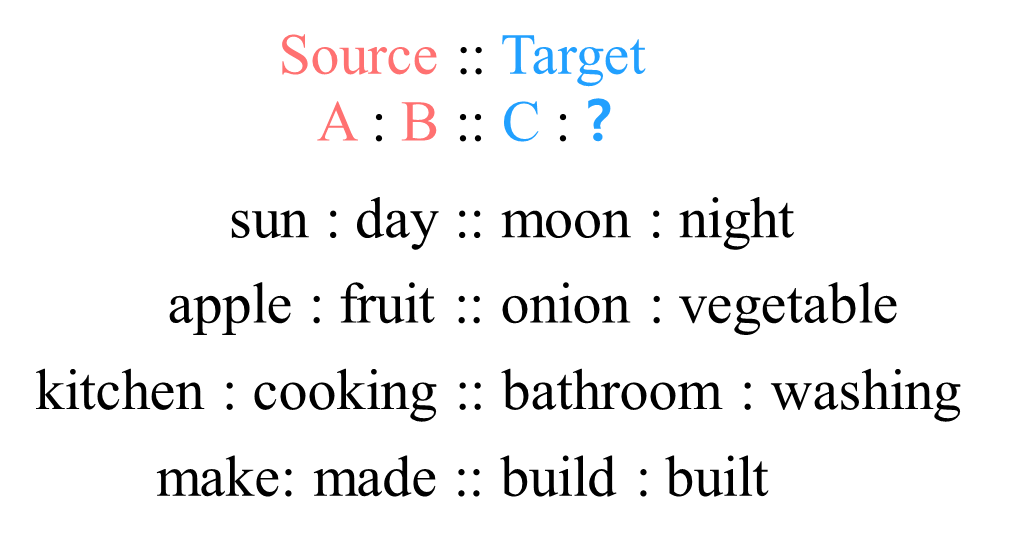}
        \caption{Proportional Analogy with words} \label{analogy_concept_word}
     \end{subfigure}
     \begin{subfigure}[b]{0.29\textwidth}
         \centering
        \includegraphics[width=0.9\linewidth]{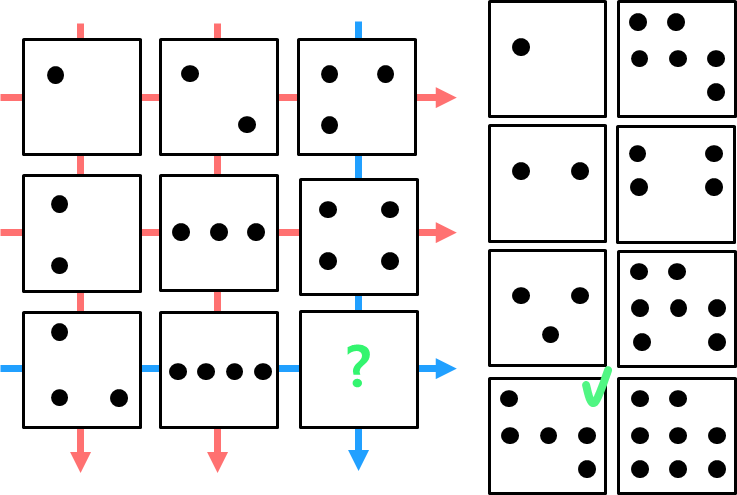}
        \vspace{0.1cm}
        \caption{Progressive Matrices} \label{analogy_concept_raven}
     \end{subfigure}
     %\hfill
     \begin{subfigure}[b]{0.32\textwidth}
         \centering
        \includegraphics[width=0.9\linewidth]{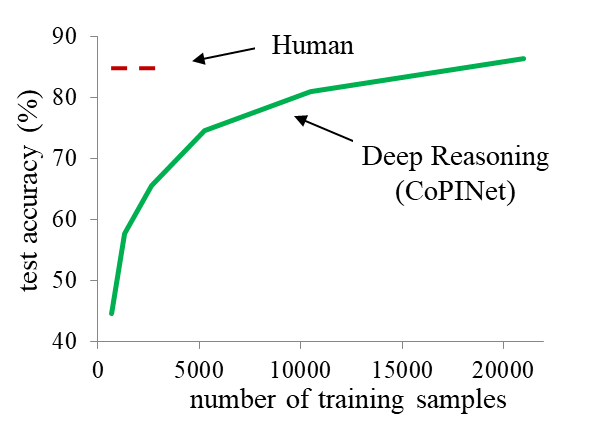}
        \vspace{-0.1cm}
        \caption{Deep visual reasoning ability} \label{analogy_human}
     \end{subfigure}
     \caption{Analogy can be explained with (a) simple proportional analogies (e.g., word analogies) and (b) visuospatial analogies (e.g., Raven's progressive matrices~\cite{penrose_raven_1936}).  (c) the state-of-the-art deep reasoning network~\cite{NIPS2019_copinet} shows dramatically degrading performance if the number of training samples is decreasing. } 
\end{figure}

The \emph{visual reasoning} task proposed in recent works~\cite{Barrett2018wren, zhang2019raven} often involves visual puzzles such as \emph{Raven Progressive Matrices (RPM)}, where the goal is to find an implicit rule among the given images, and predict the correct image piece that will complete the puzzle (see Figure~\ref{analogy_concept_raven}). Since one should identify a common relational similarity among the visual instances with diverse attributes (shape, size, and color), solving such a visual reasoning problem requires reasoning skills which might help to take a step further toward
general artificial intelligence. % systems with general intelligence. 

Recently, researchers in the machine learning community have proposed specific models for visual reasoning using deep learning~\cite{Barrett2018wren, NIPS2019_copinet, NIPS2019_9570, hill2019learning}. Deep neural networks are powerful and effective predictors for a wide spectrum of tasks such as classification and language modelling, and has yielded impressive performance on them. However, while humans can solve these logical and abstract reasoning problems by observing only few samples, the state-of-the-art deep reasoning models still require large amount of training data to achieve similar performance on the same task (see Figure~\ref{analogy_human}). 

We hypothesize that such sample-efficiency comes from the flexibility of human intelligence, which can generalize well across different problems by identifying the same pattern in the two problem-pairs with \emph{analogical reasoning}. For instance, given any two pairs of relationships, "$A$ is to $B$" and "$C$ is to $D$", one can form an analogy from the morphological parallelism between the two, e.g., as ``$A$ is to $B$ as $C$ is to $D$'' or ``$A : B :: C : D$'' (Figure~\ref{analogy_concept_word}). With such an analogical relationship established between the pairs, one will be able to predict any single entity given the rest of the entities. For visual reasoning problems, we could also identify a visuospatial analogy~\cite{Leavitt2011} between two \emph{sets} of visual entities. In general, any pair of sets from which we could identify certain relational or structural similarity between the two sets can suffice as an analogy. \emph{Analogical reasoning} has been extensively investigated to understand human reasoning process in the field of cognitive science~\cite{Perception_mitchell1993} and artificial ingelligence~\cite{Hummel97distributedrepresentations, AnalogyandRelationalReasoning}. 

However, the existing deep visual reasoning models do not have ability to perform such analogical reasoning. To overcome this limitation, we propose a deep representation learning method that exploits such analogical relationship between two sets of instances, with the special focus on the visual reasoning problem. This will allow the model to learn representations that capture relational similarities between the entities, in a sample-efficient manner.

Specifically, we train a reasoning network to learn the abstract concepts using several slightly modified/augmented problems manipulated from their original ones. As the visual reasoning problem includes designed (yet implicitly embedded) rules, we replace a randomly selected panel with a noise panel of the instances within each set. We then ensure that the relation between the two differently perturbed set should have the same relation, by enforcing their representation to be as similar as possible with contrastive learning \cite{pmlr-v97-saunshi19a_contrastive, DBLP:conf/nips/DaiL17}. Furthermore, for a cross-domain problem (with different source and target domains) given training and test sets that contain the same type of visual reasoning problems, we extract the structural relationships between elements in both domains and further maximize similarity between two relations via contrastive learning.
 
The contribution of our work is as follows:
\begin{itemize}
    \item We propose a simple but novel analogical learning method that can capture the relational similarity between two different sets of examples. 
    \item We validate our model on the visual reasoning task with RPM dataset, and show that it significantly outperforms existing baselines, with a larger margin under limited training samples or/and for test examples with unseen attributes.
    \item We propose a way to manipulate the structured set of training samples and use it to improve the generalization performance within an analogy point of view. 
\end{itemize}
We also believe that our proposed method has potential to be generalized for other applications.

\section{Preliminaries}
\subsection{Analogy and analogical reasoning}
\emph{Analogies}, which are defined on a pair of sets with the same structures or relation among the elements, are widely used as important heuristics to make new discoveries~\cite{sep-reasoning-analogy}, as they help obtain new insights and formulate solutions via analogical reasoning. \emph{Analogical reasoning}, which reasons about the new problem by drawing analogies between two problems, has been treated as an important topic for AI research. 

In general, if there are common characteristics between two objects, 
an analogical argument can be used as follows~\cite{sep-reasoning-analogy}: ``1) \textsf{S} is similar to \textsf{T} in certain (known) respects. 2) \textsf{S} has some further feature \textsc{Z}.
3) Therefore, \textsf{T} also has the feature \textsc{Z}, or some feature \textsc{Z}* similar to \textsc{Z}. 1) and 2) are premises. 3) is the conclusion of the argument.'' An analogy is a one-to-one mapping between \textsf{S} and \textsf{T} regarding objects, properties, relations and functions~\cite{sep-reasoning-analogy}. Not all of the items in \textsf{S} are required to correspond to that of \textsf{T}. Hence, in practice, the analogy only identifies selective shared items.

\subsection{Visual reasoning on Raven Progressive Matrices  (RPM)} 
The task of RPM can be formally defined using an \textit{ordered} set (tuple) of context panels $X = \{x_i\}_{i=1}^{8}$ arranged spatially in 3$\times$3 matrix with a missing cell at bottom-right corner, where a reasoner is a function to find an answer $y$ from an \textit{unordered} set of choices $C = \{c_j\}_{j=1}^8$ to fill the missing cell then make complete analogy matrix~\cite{NIPS2019_copinet}. 

RPM is a row-wise and column-wise invariant to permutation~\cite{penrose_raven_1936, ijcai2018-218, Carpenter1990WhatOI}. Thus, the commutative property between symmetric components is satisfied if an affine transform based rule is applied~\cite{ijcai2018-218}
: $f_c \circ g_r  ::  g_r  \circ f_c ~\forall c, r$   (e.g., $f_1 \circ g_2  ::  g_2  \circ f_1 $) where $f_c$ denotes column-wise transformation and $g_r$ denotes a row-wise transformation.
%as depicted in Figure~\ref{rpm}. 
The matrix can be written formally as,
$ \{\{x_i\}_{i=1}^{8};C(y)\}  := \{(f_c \circ g_r)(I)~\forall c,r;C(y)\}$ where $I \in \mathbb{R}^{h \times w}$ is a base image. While this example is a simplified form, 
%more complicated transform with diverse rules and attributes could represent more complicated problems. 
RPM problem can embed a variety of transformation (rule) and attributes~\cite{Barrett2018wren, NIPS2019_copinet}. A \textit{task} or a  problem type of RPM can be categorized based on rules and attributes applied.

\textbf{RPM problem and deep learning.} Recent advanced deep learning based reasoning models focus on the measuring relationships between the context and answer panels. The output score is calculated for those pairwise relation embedding. The final answer is chosen based on the answer panel where the highest score is shown. Calculating all possible relations using a number of combinations of context panels to extract relational representation has been conducted in Relation Network~\cite{NIPS2017_relationalreasoning} and Wild Relation Network (WReN)~\cite{Barrett2018wren}. Another way is to use a contrast effect by comparing the answer panel centered representation to the general representations which is resulted by aggregating all the context features~\cite{hill2018learning, NIPS2019_copinet}.

%These methods work using a lot of training cases to find general analogical sense.
%One of efficient way is to augment problem by defining analogical relation.
Most research works have been examined on Raven’s Progressive Matrix problem with large number of training samples which is more than 40 thousands (RAVEN dataset~\cite{zhang2019raven}) or 1 million (Procedurally Generated Matrices (PGM)~\cite{hill2018learning}) and much more than validation/test set. It is brought into a question about generalization for unseen tasks.

\textbf{Visual analogy learning.} Apart from RPM problem, there are some analogy based researches for a different purpose. Making analogy images by vector subtraction and addition in an embedding space has been investigated~\cite{NIPS2015_Analogy-Making}. 
%A query image can be transforming according to an example pair of related images. 
After accurately recognizing a visual relationship between images and then a transformed query image is generated accordingly. 
Disentangled representation for abstract reasoning has also been investigated in~\cite{NIPS2019_9570}. The authors investigated if a disentangled representation captures the salient factors of variation in samples. 
For visual recognition, an analogy-preserving visual-semantic embedding model that could both detect analogies has been developed~\cite{pmlr-v28-juhwang13}. These works provide good insights to understand embedding learning for a reasoning problem.
%By understanding of analogy preserving embedding could be a useful reference to investigate a reasoning problem.

\section{Meta-analogical contrastive learning}
One of the many reasons for the sample-inefficiency of existing deep reasoning networks may be their inefficient use of information from each single problem. To address this limitation, we note that we can extract multiple relations from a single problem. Basically, we can generate more visual reasoning problems from a given sample, by considering one of the context element as the answer, and the rest of the context elements as the context. We can then enforce this self-supervised problem to have the same relational representation as the original problem, as although we do not explicitly know the actual rule used to generate them, we know that they have the same relation. We expect this analogical learning to discover implicit relations within a problem which helps the model to better adapt to the new problem instances from the same reasoning task. 

\subsection{Analogical queries derived from support}\label{sec:support_query} From the above RPM, we assume that there is a missing sample in the context panel matrix apart from an empty answer cell. As this missing value can be estimated by neighbor panels if analogy rules (represented with transformation functions) are estimated (that is missing value immutation). As similar to inferring an answer panel using context panels, we can generate multiple validation cases by dividing the given 8 context panels by 7 sub-context panels and 1 dummy answer panel. We expect this is a simple but effective way to obtain relational embedding.
% \[
%  \begin{array}{r  r  r}
%     I~~   & \textcolor{purple}{?}    & f_2(I) \\ 
%   g_1(I) &(f_1\circ g_1)(I)    & (f_2 \circ g_1)(I)\\ 
%   g_2(I) & (f_1 \circ g_2)(I)   & \textcolor{green}{(f_2 \circ g_2)(I)}  \\
%  \end{array} \\
% :: 
% \begin{array}{r  r  r}
%     I~~         &    \textcolor{purple}{  (g_1(I))^{-1}\circ ((f_1\circ g_1)(I) )(I)}  &f_2(I)  \\ 
%     g_1(I)  &(f_1\circ g_1)(I)    & (f_2 \circ g_1)(I)\\ 
%     g_2(I)  &(f_1 \circ g_2)(I)  & \textcolor{green}{(f_2 \circ g_2)(I)} \\
%  \end{array} 
% \] where $Q_1(x_1, \textcolor{purple}{?}, f, g):: S(x_1, f, g)$. 

\begin{figure}
%\vspace{-0.7cm}
     \centering
     \begin{subfigure}[b]{0.41\textwidth}
         \centering
           \includegraphics[width=0.9\linewidth]{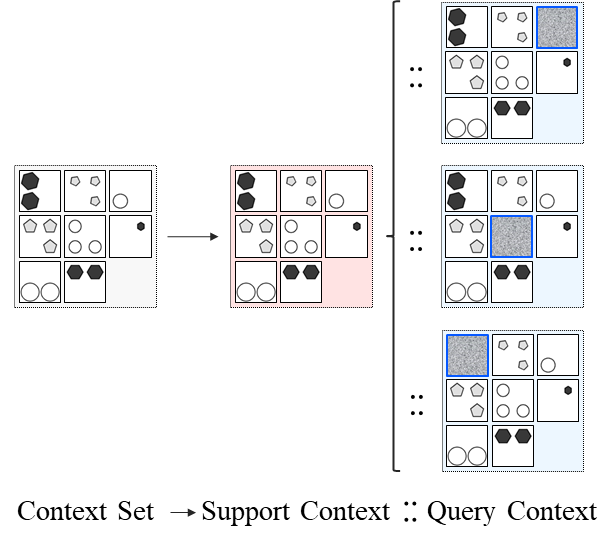}
            \vspace{-0.1cm}
           \caption{Generating multiple query context panels}  \label{fig:analogy_spt_qry}
     \end{subfigure}
     \hfill
     \begin{subfigure}[b]{0.57\textwidth}
         \centering
         \includegraphics[width=0.9\linewidth]{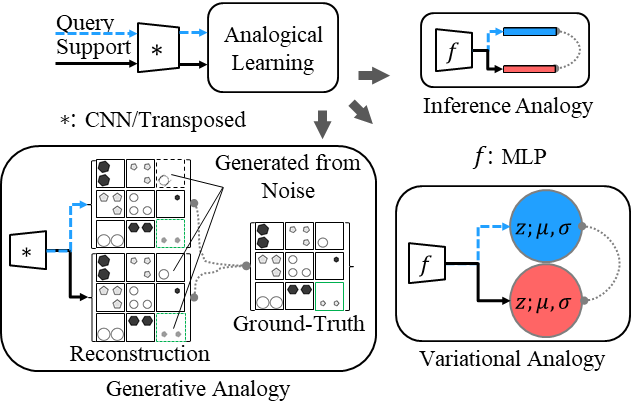}
         \caption{Analogy learning}\label{fig:analogy_loss}
     \end{subfigure}
     \caption{(a) By replacing a noise panel with a randomly selected context panel, multiple analogical pairs between support and queries can be generated. This is conducted in a contextual episode module in Figure~\ref{fig:analogy_network}. (b) Different types of analogy learning are derived using different representations such as score, latent variable, and denoised image. Those are used in learning for ``inference analogy'', ``variational analogy'', and ``generative analogy'', respectively. } 
\end{figure}

Multiple subsets derived from the original context panels ($X$) include a set of \textit{support} ($S$) with the original context panels and \textit{query} with manipulated context panels ($Q_i$) per episode. Hence, multiple analogical relation can be defined as follows: 
\begin{equation}
S:Q_1::S:Q_2::...::S:Q_k \Rightarrow  \bigcup_i\{S:Q_i\}    
\end{equation} 
where $Q_i$ is a query with a structured noise. 
An example of generating a set of support and query by inserting a noise panel into context panels is explained in Figure~\ref{fig:analogy_spt_qry}. Similar to an imputation process of missing data, implicit rules of a context set can be found by calculating a combination of those analogy relation equations. 

Let $X_S$ and $\{X_{Q_i}\}_i^{k},~\forall k$ denote the support and the query in an image level, and $S$ and $\{Q_i\}_i^{k},~\forall k$ denote the support and the query in embedding space which is transformed by encoder (e.g. CNN in Figure~\ref{fig:analogy_network}), respectively hereinafter. 

% \begin{figure}[tb]
% \centering
% \begin{minipage}[c]{0.48\textwidth}
%   \centering
%   \includegraphics[width=0.99\linewidth]{analogy_spt_qry.png}
%   \caption{By generating multiple query context panels using a noise panel, a given context set can then make multiple analogical pairs between support and queries. This explains the contextual episode.} \label{analogy_spt_qry}
% \end{minipage}
% \hfill
% \begin{minipage}[c]{0.48\textwidth}
%   \centering
%   \includegraphics[width=0.99\linewidth]{networks_anal.png}
%   \caption{Analogy learning is available with the existing learning strategy such as ``adversarial analogy learning'' with denoised and generated context panels, ``variational analogy learning'' using a latent vector of context embedding, and ``task inference analogy learning'' using task prediction. } \label{analogy_network_part} \label{fig:analogyloss}
% \end{minipage}
% \end{figure}

\subsection{Analogical learning strategy for a pair of contexts} 
Even though those multiple analogies have similar relational properties, there is no guarantee to have similar embedding for the same output. So we propose to add analogy learning to make the embedding from support and queries similar each other. Analogy between two cases can be measured with a similarity (kernel) function. Instead of the traditional gram matrix based kernel function (e.g. Gaussian kernels), 
we use a transformation based analogy kernel ($\mathcal{A}$) for effective representation learning via end-to-end training processes. 

In this paper, we use network based transformations which include a task inference encoder $e(\cdot)$, a variational encoder $v(\cdot)$, and a denoising decoder $d(\cdot)$ for a non-linear mapping of a kernel function. To define the kernel function, these transformations are combined with entropy based measures such as binary cross entropy (BCE) and Kullback–Leibler divergence (KL) so that it can be directly applied to cast inference as an optimization problem. 
%The $KL$ divergence is often referred to as the “relative entropy”. 
%an inference model, a variational encoder, and a decoder. And these are corresponded to a nonlinear mapping basis of the kernel function. Each non-linear mapping can be defined by

Formally, three analogy kernels are defined for measuring 1) similarity between task inference scores (via $e(\cdot)$) measured by BCE, 2) similarity between variational latent variables (via $v(\cdot)$) measured by KL, and 3) similarity between generative context panels ($d(\cdot)$) and ground truth context panel including correct choice panel measured by BCE. As depicted in Figure~\ref{fig:analogy_loss}, $e(\cdot)$ and $v(\cdot)$ is defined by MLP and $d(\cdot)$ by transposed CNN. While these functions can be used jointly, we use it individually to reduce complexity of overall learning. We denote individual analogy kernels as follows.

\textbf{Analogy via task inferences.} Analogy measure or similarity of task inference scores can be defined formally as follows: 
\begin{equation}
\mathcal{A}_e(S, Q_i) := -\mathbb{E}_{p(S)}{[\log{q(Q_i)}]}, \qquad \mbox{for all $i$},
\end{equation}\label{eq:task_inference} where $p(S)$ and $q(Q_i)$ are modelled using the encoder $e(\cdot;\theta)$ for $S$ and $Q_i$ respectively; and $e(\cdot;\theta)$ denotes an inference network. %= - p(e(S;\theta)) \log(p(e(Q_i; \theta))) - (1 - p(e(S; \theta))) \log(1 - p(e(Q_i;\theta)))

\textbf{Analogy via variational context latent.} Latent variables of embedding vectors can be measured their analogy as follows, 
\begin{equation}
    \mathcal{A}_v(S, Q_i) := -\mathbb{E}_{p(z|S)}{[\log{\frac{p(z|S)}{q(z|Q_i)}}]} := -\mathbb{E}_{p(z|v(S;\psi))} {[\log{\frac{p(z|v(S;\psi))}{p(z|v(Q_i;\psi))}}]}, \qquad \mbox{for all $i$},
\end{equation}
 where $v(\cdot;\psi)$ denotes a variational encoder network parametrized with $\psi$, and for availability of backpropagation, a reparameterizaiton trick is adopted as $p(z|\cdot) = \mu(\cdot) + \sigma(\cdot)\odot \epsilon$ with a mean ($\mu(\cdot)$), a variance ($\sigma(\cdot)$), and stochastic random variable $\epsilon \sim \mathcal{N}(0,1)$, activated by a sigmoid function, following variational inference.
 
\textbf{Analogy via generative contextual images.} Finally, analogy between reconstructed images of the support and the query is defined by measuring their similarity to the original context panels $X_S$ as follows,
\begin{align} 
    \mathcal{A}_d(X_S, S, Q_i)  := & -\mathbb{E}_{ X_S}{[\log{q(S, Q_i)}]}  :=-\mathbb{E}_{X_S}{[\log{q(S)}]}  -\mathbb{E}_{X_S}{[\log{q(Q_i)}]},\qquad \mbox{for all $i$},
\end{align} 
%  = - p(d(S;\pi)) \log(p(d(S;\pi))) - (1 - p(d(S;\pi))) \log(1 - p(d(S;\pi)))  - p(d(S;\pi)) \log(p(D(Q_i;\pi))) - (1 - p(d(S;\pi))) \log(1 - p(d(Q_i;\pi))),
where $X_S$ denotes the original context panel images of support $S$, $q(S)$ and $q(Q_i)$ are modelled using $d(\cdot;\pi)$, and $d(\cdot;\pi)$ denotes a decoder network parametrized with $\pi$.
%sparse imputation process is conducted during the decoding process.  This resembles metric-based meta learning.

These three analogy scores $\{\mathcal{A}_{Inf}, \mathcal{A}_{Var}, \mathcal{A}_{Gen}\}$ constitute directly an individual loss function for analogy learning. A loss for analogy learning uses one of those losses, hence,
\begin{equation}
    \mathcal{L}_{\tt analogy} = \sum_i{\mathcal{L}}(S,Q_i), %\forall i,  
\end{equation} where $\mathcal{L} \subseteq \{\mathcal{A}_{Inf}, \mathcal{A}_{Var}, \mathcal{A}_{Gen}\}$.

\begin{figure}[tb]
  \centering
  \includegraphics[width=0.99\linewidth]{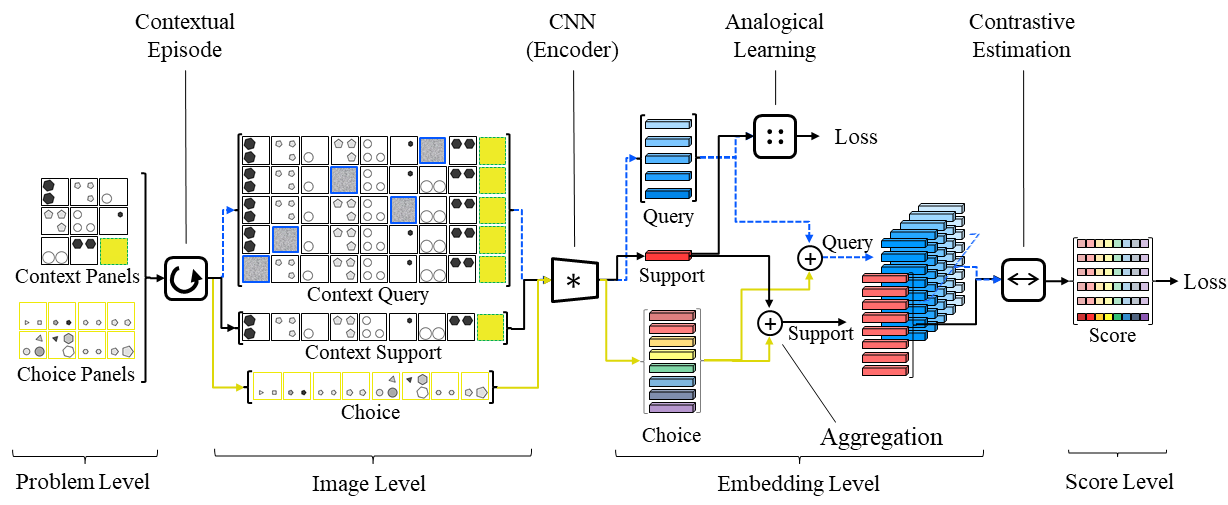}
  \caption{Our proposed networks include a contextual episode module which generate multiple analogy set between support and queries; and analogy learning module which minimize a gap between the relational characteristics extracted from those analogy set with similarity metrics. } \label{fig:analogy_network}
\end{figure}

\textbf{Contrastive estimation.}
Since the candidate choice set is an unordered list, a score for each choice panel should be independently calculated rather than associated between them. For determining of the final choice, thus, we use Noise Contrastive Estimation (NCE)~\cite{nce}. NCE is a general learning framework that estimates parameters by taking noise samples to infer the final answer instead of cross entropy which contains a normailzation factor. This framework has been used in many previous works related to detect contextual answers such as neural language modelling~\cite{nce_language2013, nce_language}, image captioning model~\cite{DBLP:conf/nips/DaiL17}, and visual reasoning~\cite{NIPS2019_copinet}.
%The basic idea is to train a softmax classifier to discriminate between samples from the data distribution and samples from some ``noise'' distribution, based on the ratio of probabilities of the sample under the model and the noise distribution~\cite{nce_language}. 
%Thus, NCE allows to fit models that are not explicitly normalized, independent of the frequency of answers appeared. In practice, by replacing noise distribution with negative sample, noise sample, general sample, or by fixing the normalizing constants instead of learning them, it achieves the similar performance to the original~\cite{nce_language}. 
We train our models using noise-contrastive estimation loss as following \cite{NIPS2019_copinet} which has proved a good performance in RPM problem. 
\begin{equation}
    \mathcal{L}_{\tt nce}(X,C,y) = \sum_i{\mathcal{L}}(f(\{X_S, X_{Q_i}\},C,y;\phi)), %\forall i,  
\end{equation} where $\mathcal{L}$ denotes NCE~\cite{NIPS2019_copinet}, and $f(\cdot;\phi)$ denotes neural networks parameterized by $\phi$, $X$ denotes the original context input, $X_S$ and $X_Q$ denote support and query panels respectively which are derived from $X$, $C$ denotes candidate answer set, and an answer label.

Overall network architecture is depicted in Figure~\ref{fig:analogy_network}. Our framework can use any existing neural networks to extract analogy embedding.

\subsection{Meta-analogical contrastive learning}   
\begin{figure}[tb]
  \centering
  \includegraphics[width=0.95\linewidth]{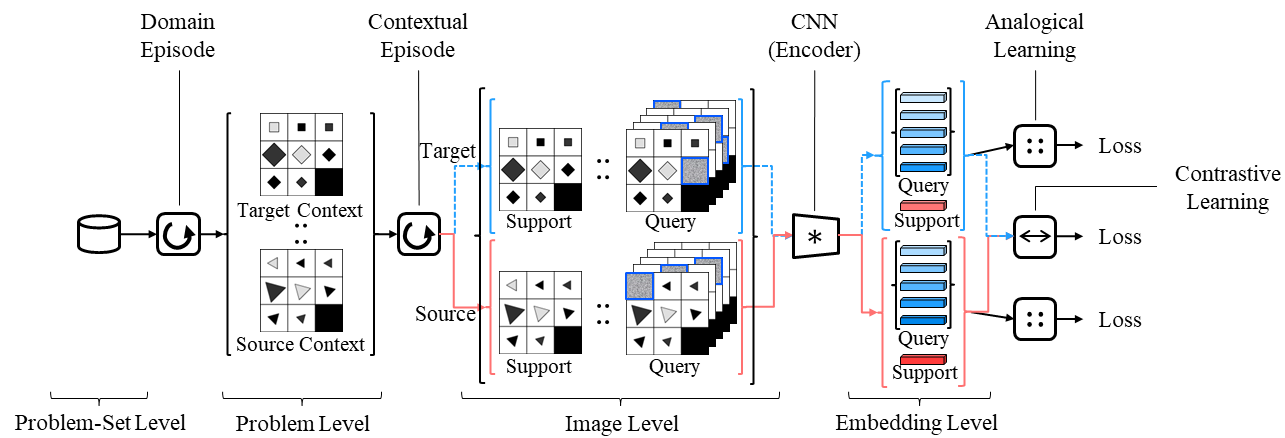}
%  \caption{Training of analogical meta learning. $x\in \mathbb{R}^{8 \times h \times w}$ is a tensor of context panels, $c \in \mathbb{R}^{8 \times h \times w}$ denotes a tensor of choice panels, and $y \in \mathbb{R}$ denotes a answer label. } 
\caption{Meta-analogical contrastive learning is based on $\{\textrm{Source context}, \textrm{Target context}\}$ composed into $\{\textrm{Support}, \textrm{Query}\}$ respectively. Domain episode module creates a domain-wise problem. Embedding from each domain can then be learned by contrastive learning. }\label{analogy_domain_meta}
\end{figure}

In the RPM problem, there are two kinds of mapping from a source domain to a target domain. First, in one problem set, `context panels' are the source domain and `subset of context panels and candidate answer panel' are the target domain as shown in Section~\ref{sec:support_query}. Then, between problems, domain mapping is found using a task category estimated based on rules and attributes. 
Source and target problems are sampled using a domain episode module (see Figure~\ref{analogy_domain_meta}). Then analogy learning can be applied according to task similarity between problems. As similar to equation~\ref{eq:task_inference}, we apply analogical contrastive learning between embedding vectors of $S$ and $Q$ from the source problem and target problem, $E_{\mathsf{S}} \in \{S, Q\}_{\mathsf{S}}$ and $E_{\mathsf{T}}  \in \{S, Q\}_{\mathsf{T}}$, respectively. 
\begin{equation}
\mathcal{L}_{\tt contrastive} = \sum_{\mathsf{S},\mathsf{T}}{\mathcal{A}_e(E_{\mathsf{S}}, E_{\mathsf{T}})} \cdot \mathrm{d}(E_{\mathsf{S}}, E_{\mathsf{T}}), 
\end{equation}\label{eq:contrastive} 
where $\mathcal{A}_e(E_{\mathsf{S}}, E_{\mathsf{T}}) := -\mathbb{E}_{p(E_{\mathsf{S}})}{[\log{q(E_{\mathsf{T}})}]}$ which is the analogy measure for a domain and $\mathrm{d}(\cdot,\cdot) \in \mathbb{R}$ denotes a soft task similarity between task scores inferred by $e(\cdot;\theta)$, (e.g., $\|e(\cdot;\theta)- e(\cdot;\theta)\|$). 

Hence, a final loss for analogy reasoning network is written as follows:
\begin{equation}
    \mathcal{L}_{\tt reasoning}(X,C,y) = \mathcal{L}_{\tt analogy} + \mathcal{L}_{\tt contrastive} + \mathcal{L}_{\tt nce}. 
\end{equation}

\section{Experiments}
\subsection{Experimental setup}
We experiment our proposed method on publicly available RPM dataset, RAVEN~\cite{zhang2019raven}. For all experiments, we train network models on the training set using ADAM optimizer. For each epoch, we evaluate the models on the validation set. Using the model shown the best validation accuracy, we evaluate the methods on the test set to report test results. All input images are resized to 80$\times$80 size. We experiment using NVIDIA Tesla V100. For the compared method, we use a publicly available implementation to reproduce results reported in~\cite{Barrett2018wren, NIPS2019_copinet}. 
More details on our experimental settings are provided in supplementary material. 
%and PGM~\cite{Barrett2018wren}

The RAVEN dataset have 70,000 problems which consists of  7 figure configurations equally distributed~\cite{NIPS2019_copinet}. The dataset is  splitted into training set (6 folds), validation set (2 folds), and test set (2 folds) per configuration. We compare our model with several simple baselines (LSTM~\cite{lstm}, CNN~\cite{hoshen2017iq}, and vanilla ResNet~\cite{HeZRS16}) and two strong baselines (WReN \cite{Barrett2018wren} and ResNet+DRT \cite{zhang2019raven}) which is reported in~\cite{NIPS2019_copinet}.

\subsection{Results}
\begin{table}[tb]
\small
  \centering
    \caption{Test accuracy ($\%$) at the best validation accuracy along different training set sizes on RAVEN dataset. The full training set has $n=42,000$ samples. The number in a bracket indicates the minibatch size used for each method. The best and second-best accuracy per each sample size (row-wise) are indicated using a bold font. }
    \begin{tabular}{r c c c c c} 
    \toprule
 & Base (CoPINet)~~ & +Analogy& +Analogy+Inf& +Analogy+Var& +Analogy+Gen\\
 %\multicolumn{5}{c}{methods}  \\ %\cmidrule{2-6}
\# samples                             & (32) & (32) & (32) & (32) & (2)\\   \midrule
  14 (0.049$\%$)   &   6.26  &  \textbf{7.68}   &   \textbf{7.48}    &     5.60  &  6.53         \\         
  35 (0.098$\%$)   &   11.07 &  13.71  &   \textbf{14.10}   &     12.35 &      \textbf{18.99}     \\         
  77 (0.195$\%$)    &   13.49 &  20.64  &   \textbf{21.31}   &     21.30    &   \textbf{27.27}     \\         
  161 (0.391$\%$)  &   21.01 &  \textbf{27.32}  &   27.27   &     25.19      &   \textbf{40.11}   \\         
  322 (0.781$\%$)  &   29.47 &  33.12  &   33.77   &     \textbf{35.54}      &  \textbf{51.96}   \\         
  651 (1.562$\%$)  &   40.34 &  48.09  &   48.30   &    \textbf{48.81}  &	\textbf{61.95}	\\ 
  \bottomrule
    \end{tabular}
    \label{tab:raven_subsample}
\end{table}
% \midrule        
%   658    &   44.48 &  -      &     -     &      -           &      \\       
%   1,316  &   57.69 &  -      &     -     &      -           &      \\       
%   2,625  &   65.55 &  -      &     -     &      -           &      \\       
%   5,250  &   74.53 &  -      &     -     &      -           &      \\       
%   10,500 &   80.92 &  -      &     -     &      -           &      \\       
%   21,000 &   86.43 &  -      &     -     &      -           &      \\  

%($n\times 0.5^{6}$), ($n\times 0.5^{7}$),($n\times 0.5^{8}$) ($n\times 0.5^{9}$)($n \times 0.5^{10}$) ($n\times 0.5^{11}$) 
\begin{table}[tb]
\centering
\small
\caption{Test accuracy ($\%$) on the RAVEN dataset. For baseline models, we report the accuracy of each model reported in~\cite{NIPS2019_copinet}. L-R denotes the Left-Right configuration, U-D Up-Down, O-IC Out-InCenter, and O-IG Out-InGrid. We denote the best results for each task with bold fonts.}
    \begin{tabular}{l c c c c c c c c}
    \toprule
    method & accuracy & Center & 2{\tiny{$\times$}}2Grid& 3{\tiny{$\times$}}3Grid& ~L-R~& ~U-D~& ~O-IC~& ~O-IG \\ \midrule
    WReN-NoTag-Aux & 17.62   &   17.66   &   29.02   &   34.67   &   7.69   &   7.89   &   12.30   &   13.94   \\
    WReN-NoTag-NoAux & 15.07   &   12.30   &   28.62   &   29.22   &   7.20   &   6.55   &   8.33   &   13.10 \\
    WReN-Tag-NoAux  &17.94   &   15.38   &   29.81   &   32.94   &   11.06   &   10.96   &   11.06   &   14.54 \\
    WReN-Tag-Aux  &33.97   &   58.38   &   38.89   &   37.70   &   21.58   &   19.74   &   38.84   &   22.57  \\
    ResNet  &53.43   &   52.82   &   41.86   &   44.29   &   58.77   &   60.16   &   63.19   &   53.12  \\  
    ResNet+DRT  &59.56   &   58.08   &   46.53   &   50.40   &   65.82   &   67.11   &   69.09   &   60.11 \\%\midrule
    CoPINet-Backbone-XE  &20.75   &   24.00   &   23.25   &   23.05   &   15.00   &   13.90   &   21.25   &   24.80 \\
    CoPINet-Contrast-XE  &86.16   &   87.25   &   71.05   &   74.45   &   97.25   &   97.05   &   93.20   &   82.90 \\
    CoPINet-Contrast-CL  &90.04   &   94.30   &   74.00   &   76.85   &   99.05   &   99.35   &   98.00   &   88.70\\ \midrule
    LSTM & 13.07 &13.19& 14.13& 13.69 &12.84& 12.35 &12.15& 12.99 \\ 
    LSTM+Analogy (ours)  & 20.70  & 21.33 & 20.40 & 20.69 &19.85 & 20.78 & 20.86 & 21.14\\      \midrule
    CNN  & 36.97   &   33.58   &   30.30   &   33.53   &   39.43   &   41.26   &   43.20   &   37.54   \\
    CNN+Analogy (ours)  & 50.37  & 44.23 & 35.47 & 39.70 & 61.33 & 53.26 & 59.09&59.34 \\\midrule
    CoPINet  & 91.42   &   95.05   &   77.45   &   78.85   &   99.10   &   99.65   &   98.50   &   91.35 \\   
    CoPINet+Analogy (ours)  & \textbf{93.06}  &  \textbf{98.00} & \textbf{78.66} & \textbf{81.77} & \textbf{99.65} & \textbf{99.70} & \textbf{99.06} & \textbf{93.48} \\     \midrule
    Human  &84.41   &   95.45   &   81.82   &   79.55   &   86.36   &   81.81   &   86.36   &   81.81 \\         
    \bottomrule
    \end{tabular}
\label{tab:raven_full}
\end{table}
 %($\uparrow  53.6\%$)  ($\uparrow 43.9\%$)($\uparrow 1.8\%$) 

\textbf{Very small (limited) training data.}
As mentioned in the introduction section, one of important tasks in analogy learning is to find an efficient learning framework which can learn something with a limited amount of samples. The state-of-the-art method for RPM dataset, CoPINet~\cite{NIPS2019_copinet}, achieved human level performance (86.43 $\% > 84.41\%$) using more than 20,000 observed problems. 
In this experiment, using less than 1.6 $\%$ of all training images (42,000 in total) which is much smaller than the number of samples conducted in~\cite{NIPS2019_copinet}, we examine generalization performance of the methods. 

As shown in Table~\ref{tab:raven_subsample}, our proposed method (+Analogy, +Inf/+Var/+Gen) achieves a significantly better performance than the state-of-the-art method, CoPINet. `+Analogy' indicates a case of analogy learning without a kernel process (+Inf/+Var/+Gen). As generative analogy learning requires decoder network (more memory), we use smaller minibatch size (2) while other compared methods used (32).

\textbf{Effect of analogical contrastive learning.} In this experiment, we show generalization performance of the proposed method using the full set of training data, by reporting the test accuracy ($\%$) on the RAVEN dataset. As shown in Table~\ref{tab:raven_full}, our proposed analogical learning strategy yields performance improvement over conventional learning methods. Particularly, with LSTM and CNN as baselines (LSTM+Analogy and CNN+Analogy) we achieve significantly enhanced performance over them with more than 50 $\%$ relative improvements. For this experiment, we do not apply NCE to examine the effect of analogical learning only. Although CoPINet has already outperformed human on this task, leaving little room for improvement, our method obtains significantly improved performance over it. 

Both WReN and ResNet+DRT highly rely on additional supervision, such as rule specifications and structural annotations. However since the RAVEN dataset provides relatively rough information (e.g. all [`Type', `Size', `Color'] parts are encoded as [1,1,1] among 9-digit rule-attributes tuple), this method seems not fit to examine. ResNet used 224 $\times$ 224 size input (identified from open code of \cite{zhang2019raven}), which means larger than input size of the RAVEN dataset (160$\times$160). Thus we did not use additional supervision for a fair comparison. 

\begin{wraptable}[12]{r}{0.45\textwidth} 
  \small
    \centering
    \vspace{-0.5cm}
    \caption{Different visual shape attributes are used among train and valid/test sets. The best accuracy is indicated using a bold font.} 
    \begin{tabular}{l c c} 
    \toprule
  ~~~~~~\multirow{2}{*}{method}       &  \multicolumn{2}{c}{minibatch size}     \\  \cmidrule{2-3}
                                &   2    & 32    \\ \midrule
  Base (CoPINet)                & 48.75  & 25.00  \\  \midrule
  ~~+ Analogy    	        & 58.75  & 51.25   \\
  ~~+ Inference Analogy    & 61.25  & 45.00  \\
  ~~+ Variational Analogy  & 56.25  & 43.75  \\
  ~~+ Generative Analogy   & \textbf{73.75}  & 43.75   \\ 
    \bottomrule
    \end{tabular}
    \label{tab:raven_cross_small}
\end{wraptable}

\textbf{Cross domains for unseen visual experience.}
Flexible intelligence can be measured by testing an unobserved problem. In this experiment, we compare the proposed method with CoPINet in terms of unobserved attributes cross training and test sets. We use training samples that contain the same type of visual reasoning problems along a training set and validation/test sets but different shapes are included differently from train (``triangle'', ``square'', and ``hexagon'') to val/test (``pentagon'' and ``circle''). 

As shown in Table~\ref{tab:raven_cross_small}, our proposed analogy learning method shows significantly improved performance (more than 50$\%$ relative improvement) compared to the baseline method. In this experiment, we use center configuration of RAVEN which is the most simple and distinctive configuration. 
When the smaller minibatch size (2) is used, it showed a better generalization performance. This might be because each problem is dissimilar each other. We used large number epochs (400) so that training loss converges. We also provide an ablation study about meta-analogical contrastive learning using pairwise samples on cross-domains (across different configurations) in the supplementary material.

\section{Conclusion}
In this paper, we proposed a framework to learn the representations for abstract relations among the elements in a given context, by solving analogical reasoning problems. We propose to exploit the \emph{analogy}, the relational similarities between two given reasoning problems. Specifically, we focused on the analogy between the set representing the original problem instance and its perturbation, as well as to another problem instance from the same task class, and then enforced the representations of the pair of instances to be as similar as possible. We further propose a generative analogy which generates a sample that preserves the analogical relationship between two problem instances in the given pair. Moreover, to allow the model to solve for unseen tasks, we further meta-learn our contrastive learning framework over pairs of samples with different attributes. We first validated our method on both visual reasoning on the full RAVEN dataset against relevant baselines including the state-of-the-art method (CoPINet), which it largely outperforms. We further validated our analogical contrastive learning framework on a challenging few-shot visual reasoning problem, on which it yields even larger improvements over CoPINet. The results of few-shot visual reasoning experiment on unseen tasks is even more impressive, as our method outperforms CoPINet by $15\%$ to $26.25\%$. While we mostly focused on visual reasoning task, our proposed analogical learning strategy may be useful for other application areas where similar relational structures could be found. 

%Furthermore, for a cross-domain problem (source and target domains), our proposed method outperformed the baseline method. Our proposed analogical learning strategy can be generalized to other application with structured data.

\clearpage
\section*{Statement of Broader Impact}
We investigated a low-shot visual reasoning problem requires reasoning skills which might help to take a step further toward general artificial intelligence in this paper. We introduce a new visual analogical learning technique. By defining multiple analogy problems with analogical contrastive learning, our method shows significantly improved generalization performance using a limited number of training samples. Furthermore, the method shows a good generalization over cross domains with different attributes. Thus, our contribution can be applied in a wide range of application in machine learning, from natural language processing to computer vision, under reasoning concepts.

Reasoning may have some ethical issues, because it is a research topic more about general intelligence which is related to human thinking. In future, for example, one can use reasoning frameworks to show the superiority of a group of people over other people. However, we do not think our work has ethical implications, because we experimented on simple and generated visual reasoning dataset (Raven's Progressive Matrices used for IQ-test) which does not include any parts to overwhelm or replace human's ability on general intelligence in the real-world. This, thus, cannot be used to prove the correctness of the reasoning problem.

%\clearpage
\bibliographystyle{unsrt}
%\bibliography{reasoning}

\clearpage
\appendix
\pagebreak
%\widetext
\begin{center}
\textbf{\large Supplementary Material: Few-shot Visual Reasoning \\ with Meta-analogical Contrastive Learning}
\end{center}

\section{Experimental results}
In this section, we examine our proposed meta-analogical contrastive learning using pairwise input samples from cross-domains. Firstly, we show an effect of contrastive learning based on relation between problems in addition to analogical learning. Then, we examine if our analogical learning is effective on analogical sampling based learning framework such as existing few-shot meta learning framework. Additionally, we show generalization performance of our proposed method across different visual domains.

%\subsection{Meta-Analogical contrastive learning}
As mentioned in the experimental section in main text, we use training samples that contain a similar type of visual reasoning problems along a training set and validation/test sets, while different shapes are included from train to val/test. 

\paragraph{Meta contrastive learning of pairwise inputs over cross domains.}
For contrastive learning on a domain level (between problems), we use estimated task scores to calculate their analogical relations. Instead of hard binary categorization ($\{0,1\}$) indicating positive or negative pairs in general contrastive metric learning~\cite{pmlr-v97-saunshi19a_contrastive}, we use a soft (task) similarity metric between problems. The soft task similarity between task scores Eq. (7) in main text, $\mathrm{d}(e(E_{\mathsf{S}}), e(E_{\mathsf{T}})) \in \mathbb{R}$, can be defined in more detail as follows,
\begin{equation}
    \mathrm{d}(e(E_{\mathsf{S})}), e(E_{\mathsf{T}}) := \textsf{CReLU}(-D+p) - \textsf{CReLU}(D-n) 
\end{equation}
where $D:=||\frac{e(E_{\mathsf{S}})}{||e(E_{\mathsf{S}})||}- \frac{e(E_{\mathsf{T}})}{||e(E_{\mathsf{T}})||}||$ denotes a normalized distance between score vectors, $e(\cdot)$ denotes the score vector of an inference encoder, $E_{\mathsf{S}}$ is a sample from a source domain, $E_{\mathsf{T}}$ from a target domain, $\textsf{CReLU}$ denotes the Rectified-linear-unit (ReLU) with clamping defined as $min(max(0, x), 1) \in [0, 1]$, and $\{p, n\}\in [0, 0.5]$ is a constraint term for positive and negative relation which aims to make samples having small distance ($<p$) closer and samples having large distance ($>n$) farther, respectively. 

As shown in Table~\ref{tab:raven_cross_small_contr}, our proposed method with meta-analogical and contrastive learning (MetaContrast) using pairwise problems (two inputs) shows improved performance compared to that of using a pointwise problem (single input). 

\begin{table}[htb]
    \centering
    \caption{Test accuracy ($\%$) at the best validation accuracy is shown, where different visual shape attributes are used among train and valid/test sets. The improved accuracy by MetaContrast (second line at each row) is indicated using an underline compared to their counterparts (first line at each row, also shown in Table 3 in main text). A bold fold indicates the best accuracy.} 
    \begin{tabular}{l c c} 
    \toprule
  ~~~~~~\multirow{2}{*}{method}         &  \multicolumn{2}{c}{minibatch size}     \\  \cmidrule{2-3}
                                        &   2    & 32     \\ \midrule
          Base (CoPINet)                & 48.75  & 25.00  \\  
          Base (CoPINet) + MetaContrast & \underline{51.25}  & \underline{48.75}  \\  \midrule
          ~~+ Analogy    	            & 58.75  & 51.25   \\
          ~~+ Analogy + MetaContrast    & \underline{61.25}  & 50.00 \\ \midrule
          ~~+ Inference Analogy         & 61.25  & 45.00  \\
          ~~+ Inference Analogy+ MetaContrast    &  \underline{62.50}   &  \underline{55.00}
  \\\midrule
          ~~+ Variational Analogy  & 56.25  & 43.75  \\
          ~~+ Variational Analogy+ MetaContrast  & \underline{62.50}  & \underline{53.75}  \\ \midrule
          ~~+ Generative Analogy   & \textbf{73.75}  & 43.75   \\  
          ~~+ Generative Analogy+ MetaContrast   &63.75 & \underline{47.50}   \\ 
    \bottomrule
    \end{tabular}
    \label{tab:raven_cross_small_contr}
\end{table}

\paragraph{Analogy sampling based meta learning.}
In this section, we experimented analogy sampling based reasoning using the given RPM problem definition (called problem category or task, as mentioned above) from RPM dataset generation. The meta information from RPM dataset includes discrete codes composed of attributes and rules. With the given \textit{problem category (task)}, a subset for learning can be sampled (via domain episode module in Figure 4 in main text). In each subset, a training sample and a test sample can have analogy relation dependent on their problem categories. In this experiment, we apply conventional meta (few-shot) learning (e.g.  Model-Agnostic Meta-Learning (MAML))~\cite{pmlr-v70-finn17a}, which consists of $K$-shot and $N$-class training samples for classification task. Here, by replacing \textit{class} with \textit{task}, $K$-shot and $N$-task reasoning framework can be defined. 
With this setting, we examined our proposed method using analogical sampling which contains different attributes (shape) but similar rule for meta-training (source) and meta-test (target) sets. This analogy sampling resembles hard task label based contrastive learning.

Here, we show analogical learning with the existing meta learning framework for fast adaptation from the source domain to the target domain. As similar to above experiments, CoPINet is used to baseline model in MAML. We use a public PyTorch based MAML code\footnote{available at~\url{https://github.com/tristandeleu/pytorch-maml}
}. We experimented this method on the same dataset used in the experiment above while meta-test using training and test samples is different from test settings in the above. Experiments are conducted with settings: 1-shot training and 1-shot test, number of epochs = 200, batchsize = 16, and number of batches = 100 without hyperparameter optimization.

As shown in Table~\ref{tab:maml_cross}, our proposed analogy learning with analogy sampling (+Analogy Learning in the table) shows significant performance improvement in terms of test accuracy at the best validation accuracy compared to analogy sampling based meta learning (Analogy Sampling in the table). Hence, this result brought a new direction of analogy sampling based analogy learning via the meta-learning framework.
\begin{table}[tb]
    \centering
        \caption{Test accuracy ($\%$) at the best validation accuracy along different number of adaptation ways (tasks). Different visual shape attributes are used among train and valid/test sets. The best accuracy is indicated using a bold font.}
    \begin{tabular}{c c c}
        \toprule
        number of ways & Analogy Sampling & +Analogy Learning (Ours)  \\
        \midrule
         2 &40.13 &59.17            \\
         3 &64.63 &82.58            \\
         4 &62.89 &90.43            \\
         5 &55.10 &89.90            \\
         6 &59.05 &\textbf{95.67}   \\
        \bottomrule
    \end{tabular}
    \label{tab:maml_cross}
\end{table}

\paragraph{Generalization across different configurations.}
We experimented our proposed method across different configurations on RAVEN dataset including: 2{$\times$}2Grid (``distribute four'', corresponding folder name in the dataset), 3{$\times$}3Grid (``distribute nine''), O-IC (``{in center single out center single}''), O-IG (``in distribute four out center single''), L-R (``left center single right center single''), and U-D (``up center single down center single''). At each configuration, while shapes can be shared, overall visual arrangements are different each other. As similar to above experiment, we use training samples with three shapes (``triangle'', ``square'', and ``hexagon'') and val/test sample with two different shapes (``pentagon'' and ``circle'') to training for a generalization purpose.

To show generalization reasoning performance of source domains on target domains, our model is trained using samples from ``center'' configuration (training) and then tested samples from other six configurations. As shown in Table~\ref{tab:cross_configure}, our proposed method (``+Analogy'') shows significantly improved generalization performance on cross domain problems.
                         
\begin{table}[tb]
    \centering
    \caption{Test accuracy ($\%$) at the best validation accuracy along different target configurations (domain) from ``center'' configuration to measure generalization performance on cross domain problems.}
    \begin{tabular}{c c c}
    \toprule
        Target Domain & Base (CoPINet) & Analogy (Ours) \\     
        \midrule
        2{{$\times$}}2Grid & 23.75 & 38.75 \\           
        3{{$\times$}}3Grid & 18.75 & 36.25 \\           
        O-IC & 30.00 & 31.25 \\           
        O-IG & 26.25 & 48.75 \\           
        L-R & 18.75 & 25.00  \\          
        U-D & 26.25 & 26.25 \\     
    \bottomrule
    \end{tabular}
    \label{tab:cross_configure}
\end{table}

\section{Experimental settings}
\paragraph{Implementation.}
For reasoning experiments, we used PyTorch code based implementation to reproduce the baseline methods (LSTM~\cite{lstm}, CNN~\cite{hoshen2017iq}, and vanilla ResNet~\cite{HeZRS16}) and strong baselines (WReN \cite{Barrett2018wren}, ResNet+DRT \cite{zhang2019raven}, and CoPINet) based on publicly available PyTorch code\footnote{available at~\url{https://github.com/Fen9/WReN} and \url{https://github.com/WellyZhang/CoPINet}
}. For dataset generation, we used publicly available RAVEN PyTorch code\footnote{available at~\url{https://github.com/WellyZhang/RAVEN}
} in the experiment ``Cross domains for unseen visual experience''.

\paragraph{Hyper-parameter settings.}
All models were trained using the Adam optimiser, with exponential decay rate parameters $\beta_1 = 0.9$, $\beta_2 = 0.999$, and $\epsilon=1\mathrm{e}{-8}$. We fixed the random seed (`$12345$', following open codes of the baseline method), and learning rate $1\mathrm{e}{-4}$. An input size of image panels are resized from $80 \times 80$ to $160 \times 160$. Images are normalized into $[-1, 1]$. Any other transformation or data augmentation techniques are not used.

\paragraph{Model details.}
Here, we provide details for all our models.

\textbf{1) Task inferences.} 
An architecture of the task inference encoder is defined in detail as follows,  
\[ e(\cdot) := [\textsf{MLP}_1, \textsf{reshape}, \textsf{softmax}, \textsf{MLP}_2, \textsf{sum}] 
\] where
\begin{itemize}
    \item $\textsf{MLP}$ consists of an inner-product layer, 
    $\textsf{MLP}_1 \in \mathbb{R}^{(d \times (a\cdot r))}$, 
    $\textsf{MLP}_2 \in \mathbb{R}^{(r \times 64)}$, 
    \item $d$ denotes a dimensionality of a (feature) vector before a classifier, 
    \item $a$ denotes the predefined number of attributes,
    $r$ number of rules, ($a=10$, $r=6$)
    \item $\textsf{reshape}$ makes the vector size from ($\textsf{batchsize} \times (a\cdot r)$) to ($\textsf{batchsize} \cdot a \times r$),
    \item and $\textsf{sum}$ make summation along $a$ to make a feature vector as ($\textsf{batchsize} \times r$). 
\end{itemize}

\textbf{2) Variational context latent.} We provide an architecture of the variational encoder in detail as follow,  
\[ v(\cdot) := [\textsf{MLP}_{\mu}, \textsf{MLP}_{\sigma}]
\]
where $\textsf{MLP}$ consists of an inner-product layer, $\textsf{MLP}_{\mu} \in \mathbb{R}^{(d \times p)}$ for the mean latent vector, $\textsf{MLP}_{\sigma} \in \mathbb{R}^{(d \times p)}$ for the standard-deviation latent vector, $d$ denotes a feature dimension before a classifier, and $p =256$.

\textbf{3) Generative contextual images.} We provide an architecture of the denoising decoder for context images generation in detail.  
 \[ 
 d(\cdot) := [\textsf{ConvT}_1, \textsf{BN}, \textsf{DeResCNN}_1, \textsf{ConvT}_2, \textsf{BN}, \textsf{concat}([\textsf{DeCNN}_1, \textsf{DeCNN}_2]), 
 \]
  \[ 
 \textsf{ConvT}_3,\textsf{upsample}, \textsf{reshape}]
 \]where 
 \begin{itemize}
     \item an input size is ($\textsf{batchsize}\cdot9\times64\times20\times20$),
     \item $\textsf{ConvT}_1$ denotes a \textit{ConvTranspose2d} layer (which is a transposed version a 2D convolutional layer implemented in PyTorch) parameterized with [input-channel: 64, output-channel: 128, kernesize: 3, stride: 2, padding: 1, bias: False] in PyTorch,
     \item  $\textsf{BN}$ denotes the batch normalization,
     \item ResNet block with transposed convludtional layers, 
    \[
    \textsf{DeResCNN}_1 : = [\textsf{ConvT}_4, \textsf{BN}, \textsf{ReLU}, \textsf{identity} + [\textsf{ConvT}_5, \textsf{BN}], \textsf{ReLU}]
    \] with $\textsf{ConvT}_4$ (\textit{ConvTranspose2d}) with [input-channel 128, output-channel: 128, kernesize: 3, stride: 1, padding: 1, bias: False], $\textsf{ConvT}_5$ (\textit{ConvTranspose2d}) with [input-channel: 128, output-channel: 128, kernelsize: 3, stride: 2, padding: 1, bias: False], 
    
     \item $\textsf{concat}$ denotes a concatenation,
     \item  $\textsf{ConvT}_3$ (\textit{ConvTranspose2d}) with [input-channel: 64, output-channel: 1, kernelsize: 5, stride:2, padding:0, bias: False],
     \item  $\textsf{upsample}$ makes an output ($80 \times 80$) size, 
     \item \textsf{reshape} makes ($\textsf{batchsize}\cdot9\times1\times80 \times 80$) to ($\textsf{batchsize}\times9\times80 \times 80$),
    \item A transposed convolutional layer for column-wise computation,
    \[
    \textsf{DeCNN}_1 := [\textsf{ConvT}_6, \textsf{BN}, \textsf{ReLU}]
    \]
with $\textsf{ConvT}_6$ (\textit{ConvTranspose2d}) with [input-channel: 64, output-channel: 32, kernesize: 3, stride: 2, padding: 0, bias: False], and
\item A transposed convolutional layer for row-wise computation,
    \[
    \textsf{DeCNN}_2 := [\textsf{ConvT}_7, \textsf{BN}, \textsf{ReLU}]
    \]
    with $\textsf{ConvT}_7$ (\textit{ConvTranspose2d}) with [input-channel: 64, output-channel: 32, kernesize: 3, stride: 2, padding: 0, bias: False].
 \end{itemize}

\end{document}